# Improved Surrogate Modeling of Fluid Dynamics with Physics-Informed Neural Networks

*Jian Cheng Wong[1], Chinchun Ooi[1*], Pao-Hsiung Chiu[1], My Ha Dao[1]*

[1]Department of Fluid Dynamics, Institute of High Performance Computing, 1 Fusionopolis Way, #16-16, Connexis North, Singapore 138632.
*Corresponding author

Corresponding author telephone number: +65 6419 1503
Corresponding author fax number: +65 6467 0200
Corresponding author email: ooicc@ihpc.a-star.edu.sg

# Abstract

Physics-Informed Neural Networks (PINNs) have recently shown great promise as a way of incorporating physics-based domain knowledge, including fundamental governing equations, into neural network models for many complex engineering systems. They have been particularly effective in the area of inverse problems, where boundary conditions may be ill-defined, and data-absent scenarios, where typical supervised learning approaches will fail. Here, we further explore the use of this modeling methodology to surrogate modeling of a fluid dynamical system, and demonstrate additional undiscussed and interesting advantages of such a modeling methodology over conventional data-driven approaches: 1) improving the model's predictive performance even with incomplete description of the underlying physics; 2) improving the robustness of the model to noise in the dataset; 3) reduced effort to convergence during optimization for a new, previously unseen scenario by transfer optimization of a pre-existing model. Hence, we noticed the inclusion of a physics-based regularization term can substantially improve the equivalent data-driven surrogate model in many substantive ways, including an order of magnitude improvement in test error when the dataset is very noisy, and a 2-3x improvement when only partial physics is included. In addition, we propose a novel transfer optimization scheme for use in such surrogate modeling scenarios and demonstrate an approximately 3x improvement in speed to convergence and an order of magnitude improvement in predictive performance over conventional Xavier initialization for training of new scenarios.

## 1. Introduction

In recent years, deep learning models have found great success across numerous application areas, especially in computer vision and natural language processing [1]. A major bottleneck in the construction of deep learning models for science and engineering systems is the fact that data collection is typically very difficult, whether through computationally expensive models or financially expensive real-world experiments. This is exacerbated by the potential need for very specific expertise to generate the data, for example in aerospace engineering, in contrast to more intuitive domains such as computer vision. It has thus been hypothesized that encoding physics-based knowledge or physics engines can be a more data-efficient way to create a deep learning model for systems with known governing physics, and there has been much work in this field in recent years [2-4].

There are many strategies to the incorporation of knowledge or physics into machine learning models, such as in the design of physics or problem-specific model architecture or physics-guided feature selection [5, 6]. In particular, the approach of incorporating physics-based knowledge or constraints into the model as a form of regularization has yielded very promising results. This ranged from the ability to provide physically-consistent solutions to challenging inverse problems across different domains such as fluid dynamics and electromagnetics [7-13], inverse discovery of hidden physics [14-16], and in the area of surrogate modeling for purposes such as uncertainty quantification and design optimization [3, 17-22]. The inclusion of physics-based knowledge as prior information during the training process via the introduction of additional regularization or constraints has also been hypothesized to be effective for remedying data-sparse or even data-absent scenarios [3, 17, 18, 23]. The nomenclature in literature is somewhat variable, with the most common names being physics-informed, physics-guided, physics-constrained or theory-guided neural networks. In general, the constraints embodied by the physics-based regularization can range from governing equations in the form of differential equations to simpler empirical or physical laws such as physical equations of state. Also, in the extreme case where no data is provided to the neural network, the physics-informed or physics-guided neural network essentially functions like a differential equation solver, similar in philosophy to early work by Lee et al. and Lagaris et al., and has been demonstrated for complex geometries, arbitrary boundary conditions and high-dimensional problems [21, 24-30].

In this work, we explore the previously described use of a physics-based regularization term in the loss function in the training and construction of surrogate models for a fluid dynamical system, and explore the potential benefits of this methodology beyond improved predictive error for data-sparse or data-absent scenarios. Specifically, we note there are two common concerns over the use of purely data-driven surrogate models, especially in the context of sparse datasets, and utilize a fluid dynamical system as a case study for exploring whether this methodology is beneficial to addressing these issues. Namely, these issues are the applicability of purely data-driven surrogate models to out-of-dataset extrapolation and the possibility of overfitting to dataset noise where data is sparse.

In that context, we first show in accordance with prior work in literature that the inclusion of physics-based regularization reduces prediction error in the final trained models obtained from a single dataset. Essentially, as the models become consistent with underlying governing physics, the model's predictions are improved. More importantly, we observe in our numerical experiments that even the inclusion of incomplete governing physics in the loss function can still lead to a better trained model.

In addition, while the training dataset used originates from numerical simulations, and is hence inherently noise-free, noise in the dataset, especially for measurement data, can be a concern in reality. While there have been other developments in adapting PINNs to handle noise via Bayesian formulations, we choose to test the ability of the base physics-based regularization to improve the neural network surrogate model in this work [31]. Hence, we evaluate the model in an extreme scenario where the data is both sparse and noisy, which might be representative of many engineering systems where sensor systems might be expensive and difficult to deploy, and demonstrate that inclusion of physics can indeed improve the neural network's robustness to noise in the data.

Lastly, while the inclusion of physics-based constraints provides a means of data-free learning for out-of-dataset extrapolation in engineering problems, we note that this physics-based training process can be slow under typical SGD-based approaches, although there have been approaches suggested to accelerate this convergence process [32, 33]. Hence, we propose building on the related concepts of transfer optimization and few-shot or zero-shot learning as an alternate approach, and demonstrate that the use of related data or data-physics models for initialization can help to accelerate physics-based training and improve predictions for extrapolations out of the starting dataset's parameter space [34-37].

## 2. Methods

### 2.1 Fluid dynamic model

For this work, we simulated the canonical case of a 2-D lid-driven cavity, which is a common benchmarking problem in computational fluid dynamics (CFD) [38, 39]. An incompressible CFD solver was used to solve for velocity in the x and y axis, and pressure (henceforth referred to as u, v and p). The velocity and pressure fields were solved for 3 different mesh resolutions, on a uniform 32 × 32 grid, a uniform 96 × 96 grid and a uniform 1024 × 1024 grid. The two sets of 32 × 32 and 96 × 96 data points were assumed to be synthetic "experimental" points for different experiments in this work, while the set of 1024 × 1024 grid points was used as ground truth for evaluation of test error.

In fluid dynamics, the non-dimensional Reynolds number (Re) is commonly used to characterize the flow, with Re defined as:

$$Re = \frac{\rho U L}{\mu} \quad (1)$$

where U and L refer to the characteristic velocity and length scale in the problem, and ρ and μ represent the density and viscosity of the fluid respectively.

For this work, the physical fluid domain simulated is of length 1 x 1 unit, hence L is 1. Density (ρ) is assumed to be constant and of magnitude 1. Five Re scenarios are simulated via our numerical simulations and used for modeling in this work, comprising $Re = \{50, 100, 150, 200, 300\}$. The different Re scenarios were created by varying their corresponding values of viscosity (μ) as given by Eq 1. Hence, values of $\mu = \{0.02, 0.01, 0.00667, 0.005, 0.00333\}$ were used for the $Re = \{50, 100, 150, 200, 300\}$ cases respectively. Two velocity Dirichlet boundary conditions for the top surface were used, with the first being a constant magnitude of 1, and the second being a polynomial function of form $u_{top} = 16x^2(1-x)^2$, while the other three surfaces were stationary surfaces ($u = 0$ and $v = 0$). Nonetheless, in both scenarios, the characteristic velocity scale (U) remains of magnitude 1. Sample contour plots of u, v and p for $Re = 100$ are provided in Figure 1, illustrating the typical fluid features expected for this canonical fluid dynamic case.

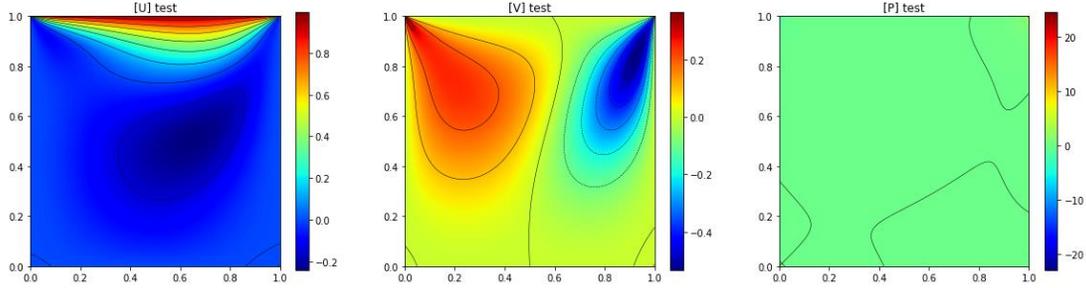
Figure 1. Representative contour plots of u, v, p for Re = 100 as obtained from numerical simulations.

## 2.2 Physics-informed neural network model

A fully-connected feed-forward neural network is used for all experiments in this work. The networks for u, v, p consist of 6 hidden layers, with 100 nodes per layer, as depicted in Figure 2. The first three layers are shared for u, v and p, while the subsequent three layers are kept distinct for each flow variable. In addition, all layers use the 'tanh' activation function, in line with prior PINN literature [7]. All networks are trained with ADAM, and training is terminated upon reaching either a pre-determined number of epochs or loss criteria.

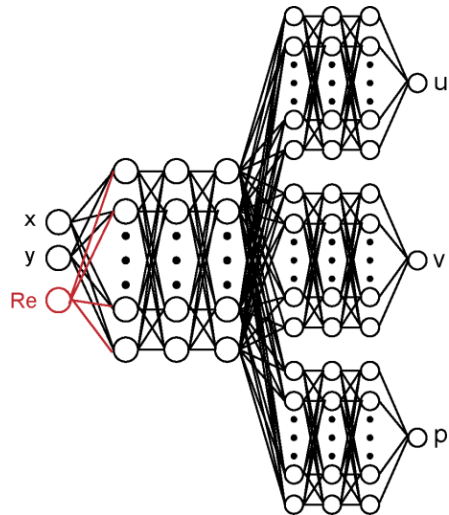
Figure 2. Neural network architecture used for this work. The node in red (Re) is added for the models involving multiple Re. There are 100 neurons per hidden layer, and the activation function is tanh.

Two separate architectures were used, with the first being used to predict the u, v, p values within the domain for a single Re, while the second was used to predict the u, v, p values across different Re (i.e. a meta-model across different values of Re). Correspondingly, Re is used as an input parameter for the second architecture but not the first, as indicated by the red portion of Figure 2. Regardless, all other parameters, including the number of nodes, number of hidden layers, and choice of activation function are kept constant across both architectures.

In this particular work, we consider the following partial differential equations (PDEs) for incompressible steady-state fluid systems to be physical laws that must be satisfied across the whole fluid domain $\Omega$ and at the domain boundary $\Omega_\Gamma$. As these equations represent the conservation of mass and momentum, these equations are representative of physics-based laws that can be implemented in general problems that should be valid.

$$\nabla \cdot \mathbf{u} = 0 \tag{2}$$

$$\mathbf{u} \cdot \nabla \mathbf{u} + \frac{1}{\rho}\nabla p - \frac{\mu}{\rho}\nabla^2 \mathbf{u} = 0 \tag{3}$$

$$\mathbf{u} = \mathbf{u}_\Gamma \text{ at } \Omega_\Gamma \tag{4}$$

Eq. 2 is a statement of conservation of mass, and is also called the continuity equation, while Eq. 3 is a statement of conservation of momentum in the x and y direction respectively for a 2-dimensional scenario as studied in this work, and is also referred to as the momentum equation. $\mathbf{u} = (u, v)$ refers to the two components of velocity in Cartesian coordinate $\mathbf{x} = (x, y)$, while p represents pressure. $\rho$ and $\mu$ represent the density and viscosity of the fluid respectively, as is consistent with Eq. 1. Eq. 4 is a statement of the relevant boundary conditions that must be satisfied for the particular system under the relevant governing equations.

Mean squared error (MSE) is used to compute the training loss function for optimization in this work, and the training loss function is defined with the following form:

$$Loss = MSE_{data} + \lambda * MSE_{PDE} \tag{5}$$

The inclusion of PDE loss term, $MSE_{PDE}$, is the hallmark of physics-informed neural networks, and is computed from the residuals of the continuity and momentum equations in Eq. 2 and Eq. 3 via automatic differentiation while the MSE on training data, $MSE_{data}$, is computed as:

$$MSE_{data} = \frac{1}{n}\sum (u_d - u_p)^2 + (v_d - v_p)^2 + (P_d - P_p)^2 \qquad (6)$$

The subscripts d and p in Eq. 6 refer to the values previously obtained from our numerical simulations (assumed ground truth) and the neural network prediction respectively.

For each experiment, the neural network's parameters are randomly initialized with Xavier normal initialization [40], and trained across 10 replicates. MSE on test data is calculated and used as the metric for model assessment based on ground truth from numerical simulations. The distribution of MSE from the predictions across 10 replicate experiments are assessed for various case studies and discussed in the Results section.

## 3. Results

### 3.1 Case studies

While benefits from the inclusion of physics into neural networks in the context of surrogate modeling have been proposed, others have noted the lack of improvement relative to purely data-driven models in their own work when data is plentiful [3, 17]. Even setting aside potential improvements in data augmentation, we hypothesize that the inclusion of physics-based regularization could be beneficial in the following ways: 1) improvement in the model predictive performance due to the additional constraints on the optimization, when the system's governing physics is fully known; 2) improvement in the model predictive performance due to the additional constraints on the optimization, when the system's governing physics is only partially known; 3) robustness to noise in sparse datasets due to the regularization effect of the physics; and 4) improved initialization for extrapolation to new scenarios without data. Hence, four sets of numerical experiments are conducted to validate these proposed benefits on a canonical fluid dynamic test case.

In the first scenario, the neural network is trained for different values of λ as per Eq. 5, where the prediction task is to model the velocity and pressure fields within a single fluid domain for Re = 100. In this instance, data points are provided on a uniform 96 × 96 grid, while test errors are evaluated on a denser, uniform 1024 × 1024 grid. Hence, the objective in this task is to evaluate the ability of the neural network to produce higher resolution predictions based on lower resolution data, which is an area of intense interest for super-resolution imaging [41]. This is also conceptually similar to the problem of inverse modelling the entire domain from a limited set of experimental (sensor) data commonly seen in literature [7, 8].

In the second scenario, a different neural network is trained to predict the velocity and pressure fields for $Re = \{50, 100, 150, 200\}$. In this instance, the training data provided is for a set of 96 × 96 points for $Re = \{50, 150\}$, while test errors are evaluated for a 1024 × 1024 set of points across $Re = \{50, 100, 150, 200\}$. Hence, this evaluates the ability of the model to provide both super-resolution interpolation within a single domain where coarse resolution data is provided, and the significantly more difficult task of interpolating and extrapolating to a parameter space, $Re = \{100, 200\}$, where no data is available.

In the third scenario, the utility of incorporating physics into the neural network model is further tested for a situation whereby the data is both sparse and noisy. In this instance, the training data is randomly sub-sampled from the set of 96 × 96 points for $Re = \{50, 150\}$, and a random fluctuation of 1% amplitude is introduced to the individual points. Test errors are similarly evaluated for a 1024 × 1024 set of points across $Re = \{50, 100, 150, 200\}$.

Finally, in the fourth scenario, the impact of physics-based regularization on the optimization process is also evaluated, especially for the purposes of extrapolation to flow conditions that were not seen by the model before. In this instance, the training data provided is for a set of 32 ×

32 points for $Re = \{50, 100\}$, while test errors are evaluated for a $1024 \times 1024$ set of points across $Re = \{50, 100, 150, 200, 300\}$. In particular, different training strategies are evaluated for their test error and relative ease of optimization.

### 3.2 Improved model performance by increasing λ

First, we evaluate the effect of the PDE loss term on the accuracy of the trained neural network for a fluid problem with Re = 100. In total, 5 values of λ are tested, comprising 0, $10^{-3}$, $10^{-2}$, $10^{-1}$ and 1. The distribution of test errors for u, v, p from 10 experimental runs are presented in the following box plots in Figure 3.

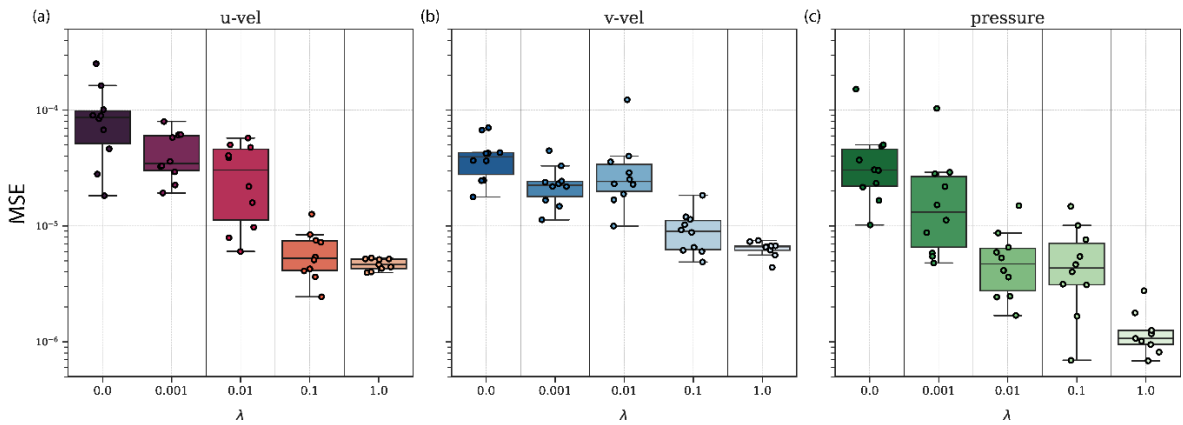

Figure 3. Boxplots of test errors observed for (a) u, (b) v, (c) p across 10 runs for 5 different values of λ.

A reduction in the test error is observed across all the field variables, which is in line with the general expectation that the provision of physics-based equations should improve the model by enforcing consistency with the physical system's underlying governing laws.

More crucially, we noticed that the mean test error across different models reduced significantly as λ was increased. This phenomenon was evident across u, v and p, with a reduction of up to an order of magnitude for λ = 1 relative to λ = 0. This suggests that the addition of a physics-based regularization term does indeed improve the neural network's effectiveness at surrogate modeling.

### 3.3 Improved prediction across multiple Re

Having observed the reduction in test error for a single case involving $Re = 100$, we further extend our work by evaluating the impact of the PDE loss term on consistency for a more complex model spanning multiple Re. As a proof-of-concept, we train the same neural network with data for the $Re = \{50, 150\}$ cases on a $96 \times 96$ grid, and use it to evaluate the test error on a $1024 \times 1024$ grid for the cases of $Re = \{50, 100, 150, 200\}$. Based on the results for the $Re = 100$ case in Section 3.1, we focus only on $\lambda = 0$ and $\lambda = 1$.

In addition, it is common in many engineering systems that we might only have incomplete knowledge of the system. Hence, we also tested the effect of providing only the continuity equation (Eq. 2), and compared it to the effect of providing both continuity and momentum equations (Eq. 2 and Eq. 3), which is a complete description of the underlying physics governing this particular fluidic system.

### 3.3.1 Improved prediction across multiple Re with only continuity equation

We incorporate only the continuity equation into the loss function for the first set of experiments. Based on the distribution across 10 runs, we noticed that there was a similar decrease in the test error when a physics-based loss term was incorporated during training ($\lambda = 1$ vs $\lambda = 0$). Results are plotted in Figure 4.

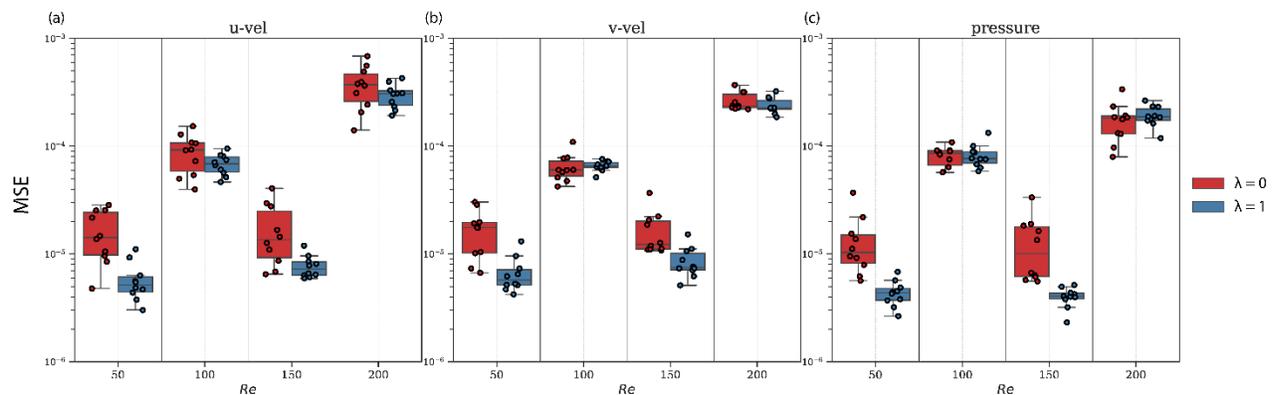

Figure 4. Boxplots of test errors observed for (a) u, (b) v, (c) P across 10 runs for 4 different Re with only the inclusion of the continuity equation during training.

We note that an improvement in prediction error does not always result across all the cases tested, but on a whole, the error was decreased for the cases with some data provided, suggesting that there was still some benefit to the inclusion of a partial description of the underlying physical system. Importantly, all runs for the cases tested reached the convergence threshold of 2 × 10$^{-4}$ during training, suggesting that training loss alone would have been insufficient for differentiating these models.

### 3.3.2 Improved prediction across multiple Re with full physics

A similar set of experiments are conducted, but with the inclusion of both the continuity and momentum equation. However, the addition of the momentum equation (Eq. 3) in the loss function impacted the convergence rate of the training process somewhat, hence a slightly higher convergence threshold of 5 × 10$^{-4}$ was used instead of 2 × 10$^{-4}$ as per the earlier experiments when only continuity equation was used. Results are presented in Figure 5.

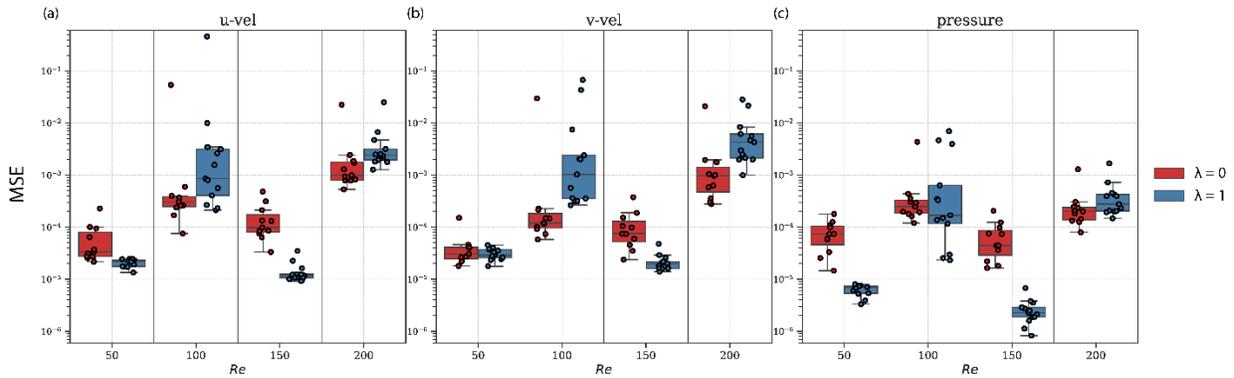

Figure 5. Boxplots of test errors observed for (a) u, (b) v, (c) P across 10 runs for 4 different Re with the inclusion of both the continuity and momentum equations during the training process.

Similarly to the case with only the inclusion of conservation of mass, the same reduction in prediction error is observed for the $Re = \{50, 150\}$ cases, which are the Re cases for which training data is provided. Somewhat surprisingly, there is no observable improvement for extrapolation to the other two cases where data was not provided in this meta-model for Re.

### 3.4 Improved predictive model for noisy and sparse data

In the earlier experiments, the models that were trained with a combination of data and physics produced a small improvement over the data-only models. However, we hypothesized that this is due to the training data being both relatively comprehensive and noise-free. Hence, as a further evaluation of the utility of including physics, additional experiments are conducted with a down-sampled dataset with noise that had been artificially added. The results are presented in Figure 6.

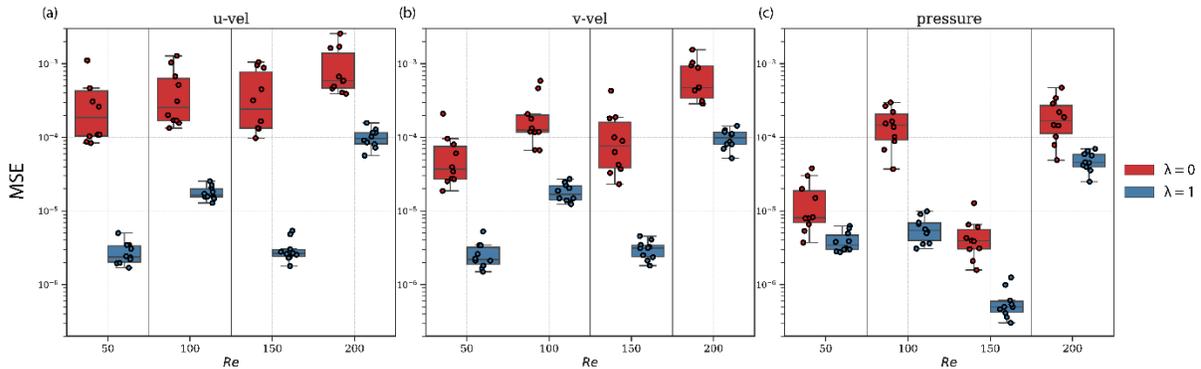

Figure 6. Boxplots of test errors obtained for (a) u, (b) v, (c) P across 10 runs for 4 different Re when the training data comprises of a down-sampled dataset with artificial random noise.

The results show that the introduction of physics into the neural network is particularly helpful when the data is both sparse and noisy, with an order of magnitude improvement in the test error with the inclusion of physics. In this instance, we note a reduction in model test error across all four Re cases.

Interestingly, the results show that the test error remains lowest for the $Re = \{50, 150\}$ scenarios where actual training data is provided, despite the addition of noise to the training data. This is true for both models regardless of whether physics was included in the neural network. This implies that the provision of data is still beneficial to the neural network and that the physics is complementary to the data. This is potentially due to the implications of having the additional data-driven term ($MSE_{data}$) in the neural network's training loss function.

### 3.5 Improved optimization with incorporation of physics

In addition to enhancements in model performance, we also explore the benefits from an optimization perspective. Some prior work in literature have demonstrated the value of physics-based loss training as a means for data-less modeling. In this work, we further explore the use of a pre-trained neural network for extrapolation to a related scenario where no data is available. Hence, we train a neural network with data from the $Re = \{50, 100\}$ scenarios, and further optimize the surrogate model to the out-of-set scenarios of $Re = \{150, 200, 300\}$.

In total, six different prediction strategies were tested across the three additional scenarios of $Re = \{150, 200, 300\}$: i) and ii) involve immediate extrapolation from a data-driven neural network (model A1) and a physics-informed neural network (model B1) that was trained for the $Re = \{50, 100\}$ scenarios; iii) and iv) include further optimization without and with physics-based constraints for the new scenario of interest from the models developed for (i) and (ii), which correspond to models A2 and B2 respectively; v) data-less training of a PINN from a random initialization (model C1). In addition, model B3 was trained as per model B2, but with the maintenance of data from the $Re = \{50, 100\}$, which might be relevant when further interpolation to prior scenarios might be of interest. The respective training strategies are summarized in Table I for clarity.

Table 1. Description of the training strategies tested

| Model | Training | | Prediction |
|---|---|---|---|
| | Step 1 | Step 2 | |
| A1 | Training with data for $Re = \{50, 100\}$ | - | Prediction for $Re = X$ where $X = \{150, 200, 300\}$ |
| A2 | Training with data for $Re = \{50, 100\}$ | Additional training with physics for $Re = X$ | |
| B1 | Training with data and physics for $Re = \{50, 100\}$ | - | |
| B2 | Training with data and physics for $Re = \{50, 100\}$ | Additional training with physics for $Re = X$ | |
| B3 | Training with data and physics for $Re = \{50, 100\}$ | Additional training with physics for $Re = X$ & data and physics for $Re = \{50, 100\}$ | |
| C1 | - | Training with physics for $Re = X$ | |

### 3.5.1 Improved optimization with incorporation of physics

In the first set of experiments, the effectiveness of models A1, A2, B1, B2 and C1 are evaluated for the scenarios of $Re = \{150, 200, 300\}$. The respective results for u, v and p are collated and plotted in Figure 7.

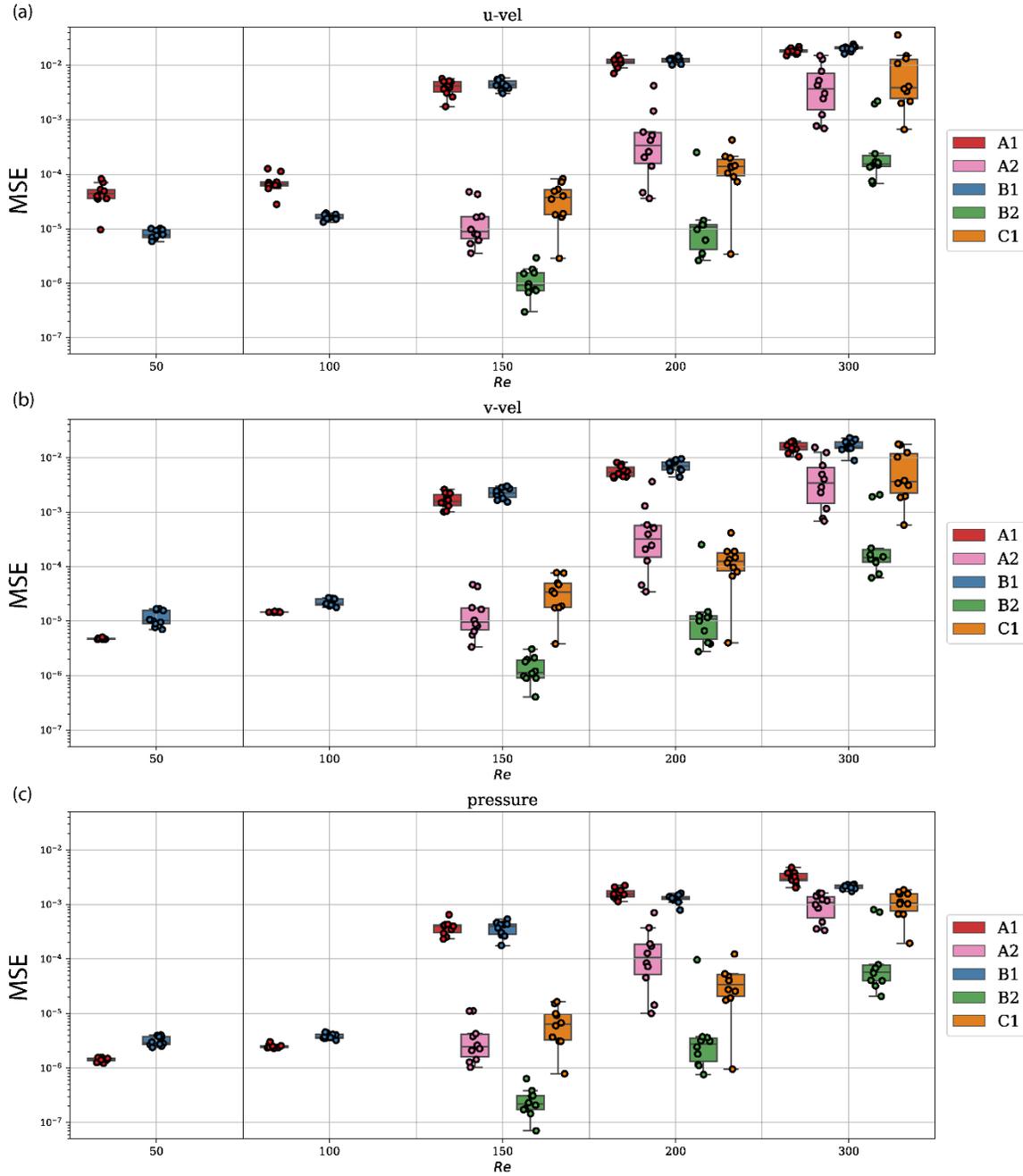

Figure 7. Boxplots of test errors obtained for (a) u, (b) v, (c) p across 10 runs for 5 different Re obtained via different model training strategies as listed in Table I.

From our experiments, the models trained for the $Re = \{50, 100\}$ scenarios did not extrapolate well to the new scenarios of $Re = \{150, 200, 300\}$ without further optimization, even when the neural network was trained with physics-based regularization (models A1 and B1). This further demonstrates that even while the inclusion of physics is anticipated to help with ensuring consistency of the neural network to the underlying governing physics, this still only extends to interpolation within a parametric space. This is also consistent with our observations in the previous sub-sections. We also note that model C1 out-performed the direct prediction from models that had been trained for $Re = \{50, 100\}$ (models A1 and B1), which is consistent with the literature showing the value of physics-based constraints as the training objective in data-less situations.

More interestingly, the model that was further trained with physics-based knowledge from a model that had been previously trained with both data and physics-based constraints for the $Re = \{50, 100\}$ scenarios (model B2) was found to out-perform the other cases, including both the model that was only trained with physics from the beginning (model C1) and the model that had been further trained from the data model (model A2). This suggests that prior training with both data and physics for an easier scenario helped the subsequent optimization converge to a better minimum during training.

The training process was also found to be accelerated by using a better initialization as obtained from prior training of the neural network with both data and physics for the $Re = \{50, 100\}$ scenarios. Representative convergence plots of the training are presented in Figure 8 for Models A2, B2 and C1.

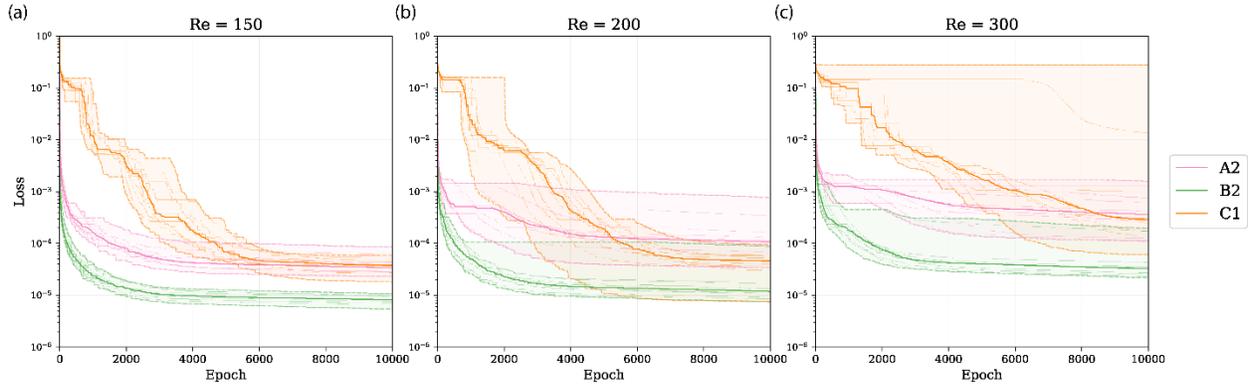

Figure 8. Aggregate line plots of the convergence curves during training for Models A2, B2 and C1 are presented for (a) Re = 150, (b) Re = 200 and (c) Re = 300.

The optimization curves plotted in Figure 8 illustrate how further optimization to a scenario involving a new Re is greatly improved both in terms of MSE error and training effort (epochs to convergence) when an existing model that had been previously trained on both data and physics for an earlier case is used as an initialization (Model B2) relative to other forms of initialization (Model A2 or C1). This improvement is particularly notable in relation to Model C1, which used a random initialization. This suggests additional utility from the optimization perspective to training a model with data and physics for an earlier, simpler scenario, and subsequently using the model for extrapolation to a different, potentially more difficult problem.

### 3.5.2  Impact of retaining old data during transfer optimization

In the process of surrogate modeling, there are situations where one wishes to retain memory of the prior data. Hence, we also present results here to demonstrate the performance of this approach, as plotted in Figure 9.

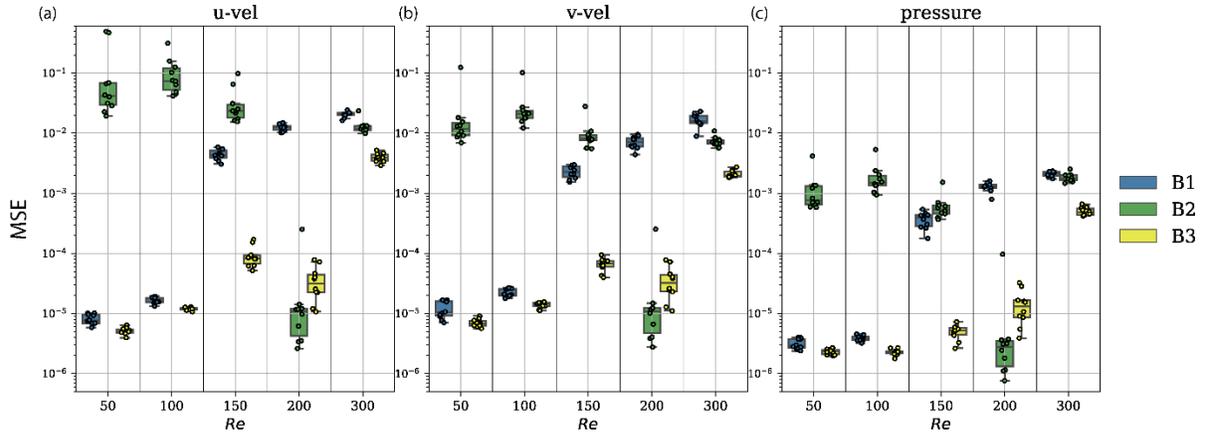

Figure 9. Boxplots of test errors obtained for (a) u, (b) v, (c) p across 10 runs for 5 different Re obtained via different model training strategies as listed in Table I.

In line with our intuition from the training process described, model B1 has good predictive performance for $Re = \{50, 100\}$ only, while model B2 has good predictive performance for $Re = 200$ only. In contrast, model B3 was trained with both the new physics constraints for $Re = \{50, 100, 200\}$ and old data for $Re = \{50, 100\}$, and hence, could both provide good predictive performance for $Re = \{50, 100, 200\}$, and interpolate well for $Re = 150$. Nonetheless, model B3 is still unable to provide a good prediction for $Re = 300$. This further corroborates our previous results whereby we note that training with physics, as in B1 and B2, does not actually provide guarantees on the model being physically consistent or better-performing for extrapolation to scenarios outside the parameter space explored during training. It is however able to provide a better interpolation for scenarios where no training occurred, but which is still within the parameter space (i.e. $Re = 150$).

## 4. Discussion

Firstly, we noticed in our experiments that the final prediction error from trained models can be minimized significantly by incorporating the underlying governing physical laws into the training process. In this instance, the dataset provided for training is already quite substantial, spanning the entire set of $96 \times 96$ points within the fluid domain, hence, the improvement in accuracy is not significant. In a subsequent set of experiments where a sparser, noisier data set was provided for training, a much more significant improvement in prediction error was indeed observed.

Interestingly, while a reduction in prediction error was consistently observed when some data was provided, this effect was much weaker for model extrapolations. As observed in Figure 5, the error for extrapolated predictions did not always improve, and the extrapolation error was typically still similarly high between the physics-informed neural network and the base neural network model. We believe that this also suggests that the use of physics-informed neural networks for surrogate modeling should be done with caution. Essentially, without sampling of the parameter space during the training process, the provision of physics does not actually have an opportunity to act on the neural network to ensure consistency of physics is similarly maintained across the extended parameter space. Hence, the physics constraints are actually still restricted within a pre-determined parameter space. Nonetheless, this is worth further study, and it is possible that different training strategies might be able to ameliorate this issue.

In addition, it was observed that the inclusion of the full set of equations, both continuity and momentum in this example, can impact the optimization process. Convergence is slowed by the addition of the PDEs into the training loss function as the additional constraints make the optimization landscape significantly more complex, especially when these constraints such as the Navier-Stokes equations are highly non-linear. While the standard ADAM optimization is used in this work, future work exploring the use of different optimization strategies for incorporating these additional physics-based constraints into the training loss function is an interesting area to explore for remedying this adverse impact on optimization.

Nonetheless, it is worth noting that even the partial inclusion of underlying physics, such as the continuity equation alone, can improve the model prediction error in our experiments. The inclusion of partial physics opens up possibilities for employing this strategy in real-world situations where even domain experts may not fully understand the physics, and has less of a detrimental impact on model convergence while retaining some of the benefits. Future work studying the trade-off between the amount of physics to include, and the additional benefits to model performance will hence be an interesting extension to this work.

Lastly, while full numerical simulations can be expensive, it is rare that surrogate models are required for engineering scenarios in the complete absence of data. Instead, there will typically be an existing dataset, although it may be sparse and may not encompass the parametric range of interest. Hence, it is preferable to utilize the existing sparse dataset and existing data models to minimize the cost of providing a high-quality new prediction. Indeed, this is the approach encapsulated by the idea of transfer learning in conventional machine learning. In acknowledging the difficulties associated with optimization of the physics-informed neural networks due to the addition of complicated physics-related constraints, especially in the no-data situations, we further show the benefits of first training a model with existing, simpler data and physics to accelerate and improve convergence for new scenarios. A potential practical approach in the future might be in utilizing the physics to facilitate further optimization from a base model that was previously trained for a generic parametric range and that had been stored.

## 5. Conclusions

We note that this is an extension on prior work suggesting that physics-based knowledge incorporation is generally expected to improve the neural network's prediction accuracy. However, there has been little additional work to our knowledge exploring other benefits of this methodology. Hence, we investigate the impact of incorporating physics-based knowledge as an extra regularization term into the training process in this work, and elucidate its benefits to the neural network beyond the typical claim of improved L2 test errors under data-scarce conditions. We uncover several interesting benefits that have not been discussed in literature, and our results provide further motivation for the incorporation of such loss terms into training and propose some interesting directions for further work.

## Acknowledgments

The authors would like to acknowledge research funding from the Agency for Science, Technology and Research (A*STAR), Singapore, under Grant No. A1820g0084.